\PassOptionsToPackage{sort&compress}{natbib}
\documentclass[preprint,12pt]{elsarticle}

\usepackage{hyperref,tikz,bbding,amssymb,pifont,adjustbox,graphicx,soul,textcomp,array,longtable,bm,multirow,xcolor,color,hyphenat, microtype}
\usepackage{soul}
\usepackage{setspace}

\graphicspath{ {./figures/} }

\usepackage[left=1in, right=1in, top=1in]{geometry}

\usepackage[T1]{fontenc}
\usepackage{makecell}

\usepackage{etoolbox}
\makeatletter
\def\ps@pprintTitle{%
 \let\@oddhead\@empty
 \let\@evenhead\@empty
 \def\@oddfoot{}%
 \let\@evenfoot\@oddfoot}
\makeatother

\newcolumntype{P}[1]{>{\centering\arraybackslash}p{#1}}
\newcolumntype{M}[1]{>{\centering\arraybackslash}m{#1}}

\usepackage{caption} 

\begin{document}
\setstretch{1.5} 

\begin{frontmatter}

\title{Eye-SpatialNet: Spatial Information Extraction from Ophthalmology Notes}

\author[mymainaddress]{Surabhi Datta}
\ead{surabhidatta92@gmail.com}
\author[mymainaddress]{Tasneem Kaochar}
\author[mymainaddress]{Hio Cheng Lam}
\author[mymainaddress]{Nelly Nwosu}
\author[mymainaddress]{Luca Giancardo}
\author[mysecondaryaddress]{Alice Z. Chuang}
\author[mysecondaryaddress]{Robert M. Feldman}
\author[mymainaddress]{Kirk Roberts\corref{mycorrespondingauthor}}
\ead{kirk.roberts@uth.tmc.edu}

\cortext[mycorrespondingauthor]{Corresponding author}

\address[mymainaddress]{McWilliams School of Biomedical Informatics, UTHealth Houston}
\address[mysecondaryaddress]{McGovern Medical School, UTHealth Houston}

\begin{abstract}
This paper focuses on the representation and automatic extraction of spatial information in ophthalmology clinical notes.
We extend our previously proposed frame semantics-based spatial representation schema, Rad-SpatialNet, to represent spatial language in ophthalmology text, resulting in the Eye-SpatialNet schema.
The spatially-grounded entities are findings, procedures, and drugs.
To accurately capture all spatial details, we add some domain-specific elements in Eye-SpatialNet.
Utilizing this representation, we annotated dataset of 600 ophthalmology notes labeled with detailed spatial and contextual information of ophthalmic entities.
The annotated dataset contains 1715 spatial triggers, 7308 findings, 2424 anatomies, and 9914 descriptors.
To automatically extract the spatial information, we employ a two-turn question answering approach based on the transformer language model BERT.
The results are promising, with F1 scores of 89.31, 74.86, and 88.47 for spatial triggers, Figure, and Ground frame elements, respectively.
This is the first work to represent and extract a wide variety of clinical information in ophthalmology.
Extracting detailed information can benefit ophthalmology applications and research targeted toward disease progression and screening.
\end{abstract}

\begin{keyword}
\texttt Information Extraction \sep Spatial Information \sep Ophthalmology \sep Natural Language Processing \sep Deep Learning \sep  Question Answering
\end{keyword}

\end{frontmatter}

\section{Introduction}
\label{sect:introduction}
Ophthalmology notes contain important clinical information about a patient's eye findings.
These findings are documented based on interpretations from imaging examinations (e.g., fundus examination), complications or outcomes associated with surgeries (e.g., cataract surgery), and experiences or symptoms shared by patients. 
Such findings are oftentimes described along with their exact eye locations as well as other contextual information such as their timing and status.
Thus, ophthalmology notes comprise of spatial relations between eye findings and their corresponding locations, and these findings are further described using different spatial characteristics such as laterality and size.
Although there has been recent advancements in using natural language processing (NLP) methods in the ophthalmology domain, they are mainly targeted for specific ocular conditions.
Some work leveraged electronic health record text data to identify conditions such as glaucoma \cite{wang2022DeepLearningApproaches},
herpes zoster ophthalmicus \cite{zheng2019UsingNaturalLanguage}, and exfoliation syndrome \cite{stein2019EvaluationAlgorithmIdentifying}, while another set of work extracted quantitative measures particularly related to visual acuity \cite{baughman2017ValidationTotalVisual,mbagwu2016CreationAccurateAlgorithm} and microbial keratitis \cite{woodward2021DevelopmentValidationNatural}.
In this work, we aim to extract more comprehensive information related to all eye findings, covering both spatial and contextual, from the ophthalmology notes.
Besides automated screening and diagnosis of various ocular conditions, identifying such detailed information can aid in applications such as automated monitoring of eye findings or diseases and cohort retrieval for retrospective epidemiological studies.
For this, we propose to extend our existing radiology spatial representation schema--Rad-SpatialNet \cite{datta2020RadSpatialNetFramebasedResourceb} to the ophthalmology domain.
We refer to this as the Eye-SpatialNet schema in this paper.
We annotate a total of $600$ ophthalmology notes following Eye-SpatialNet.
Finally, we apply an advanced deep learning-based method to automatically identify the spatial and contextual information from the notes.

\begin{figure*}[!t]
\includegraphics[width=0.9\textwidth]{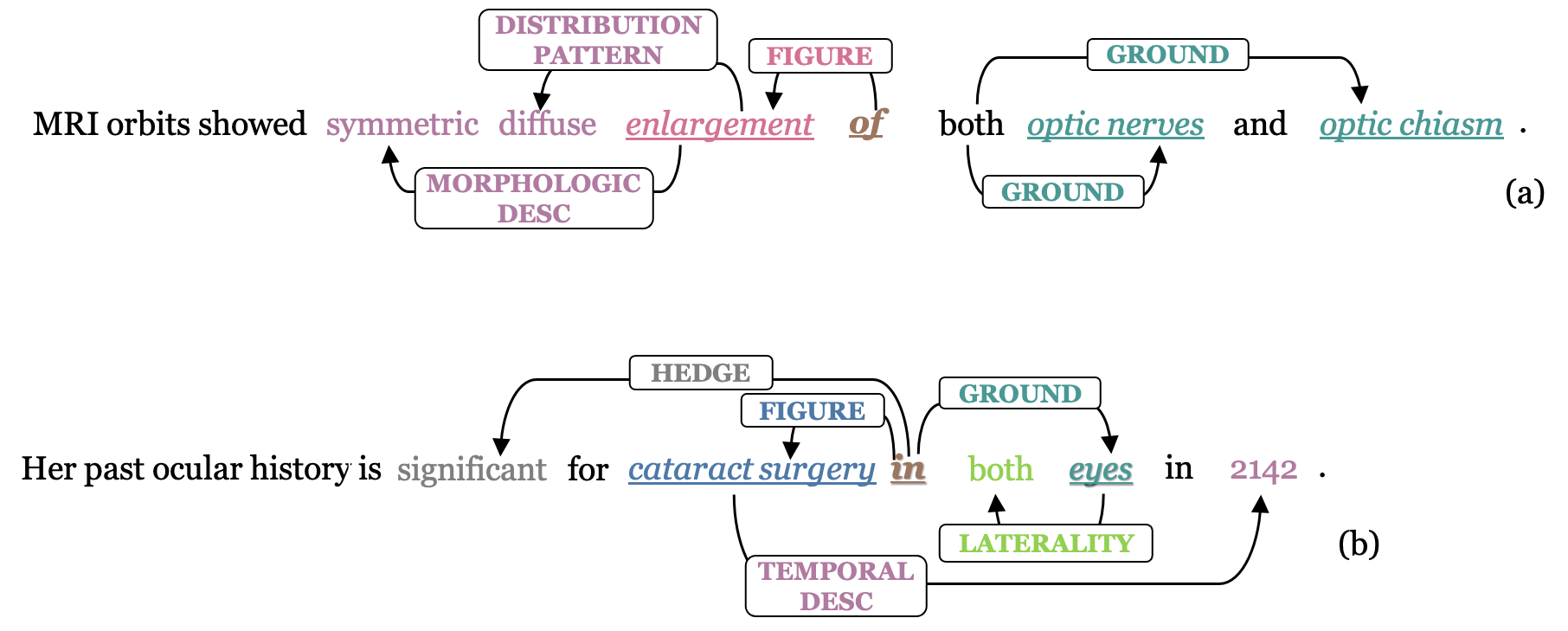}
\centering
\caption{Example sentences from ophthalmology notes showing some of the spatial frame elements covered in the Eye-SpatialNet schema. The underlined and italicized texts denote the lexical units of the frames.}
\label{fig:example}
\end{figure*}

Ophthalmologists use spatial language to describe findings interpreted from imaging techniques.
For example, in the sentence -- ``\textit{OCT of the retinal nerve fiber layer shows normal thickness in both eyes.}'', both eyes have been described using the finding \textit{normal thickness} as interpreted from an Optical Coherence Tomography examination.
Here, \textit{thickness} is spatially associated to \textit{eyes} through the preposition \textit{in}, where \textit{normal} describes the status of thickness and \textit{both} describes the laterality.
Similarly, symptoms presented by patients are also documented using spatial relations.
In the sentence -- ``\textit{She presented in [DATE] with weakness and numbness of her right eye as well as pain and vision loss in the left eye consistent with optic neuritis.}'', the findings \textit{weakness} and \textit{numbness} are spatially related to \textit{right eye} through the preposition \textit{of}, whereas \textit{pain} and \textit{vision} are linked to \textit{left eye} through \textit{in}.
Additionally, we note that the ophthalmologist also reports the potential diagnosis inferred from these findings, i.e., \textit{optic neuritis}.
\textit{[DATE]} denotes the timing associated with the findings.
Sometimes, eye procedures and drugs are also associated with anatomical locations and thus are spatially-grounded.
We capture all these important information in our Eye-SpatialNet schema.

The Eye-SpatialNet schema is based on frame semantics, where a lexical unit (LU) represents the word that invokes a frame and the participants of a frame form the frame elements (FEs).
The spatial prepositions (e.g., \textit{in}) and verbs (e.g., \textit{reveals}) constitute the lexical units whereas the associated findings (e.g., \textit{weakness}), the locations (e.g., \textit{eye}), diagnosis (e.g. \textit{optic neuritis}), and the various spatial and other descriptors (e.g., \textit{left}, \textit{normal}) constitute the frame elements.
The spatial prepositions and verbs are also referred to as spatial triggers in this paper.
Following this schema, we create a manually-annotated dataset of $600$ ophthalmology notes to represent important spatial information of clinical significance.
Two sample examples from our ophthalmology dataset are illustrated in Figure \ref{fig:example}.
Note that for (a), `Figure' and `Ground' are the spatial frame elements of the frame evoked by the spatial trigger \textit{of}, whereas `Morphologic descriptor' and `Distribution pattern'  are the spatial frame elements of the frame evoked by the finding \textit{enlargement}.
`Figure' usually refers to an entity whose location is described through a spatial trigger whereas `Ground' denotes the actual anatomical location.
In the second example (b), \textit{cataract surgery} is spatially linked to \textit{eyes} where \textit{cataract surgery} acts as the `Figure' element of the frame evoked by the spatial trigger \textit{in} and \textit{2142} (year altered for de-identification) is a descriptive frame element of the frame instantiated by the procedure \textit{cataract surgery}.
There are a total of \textcolor{black}{$1715$} spatial triggers, \textcolor{black}{$7308$} finding, and \textcolor{black}{$2424$} location phrases annotated in the dataset.
We describe the annotation process in Section~\ref{lab:annotation}.
To our knowledge, this is the first study to develop an annotated dataset with comprehensive representation schema for identifying detailed information from ophthalmology notes.

For automatic extraction of the spatial information, we adopt a two-turn question answering framework \cite{li2019EntityRelationExtractionMultiTurna,datta2022FinegrainedSpatialInformationa} based on a transformer language model, BERT \cite{devlin2019bert}.
This is inspired by previous studies demonstrating the effectiveness of framing various information extraction tasks such as named entity recognition \cite{li2020UnifiedMRCFramework}, relation extraction \cite{levy2017ZeroShotRelationExtraction}, and event extraction \cite{liu2020EventExtractionMachine} as question answering (QA) by harnessing the well-developed machine reading comprehension models.
Further, some studies \cite{li2019EntityRelationExtractionMultiTurna,li2020EventExtractionMultiturn,wang2020BiomedicalEventExtraction} investigated the formulation of relation and event extraction tasks as multi-turn QA both in the general and biomedical domain.
In this paper, we apply a two-turn QA method similar to the one proposed for radiology domain \cite{datta2022FinegrainedSpatialInformationa}, to extract the spatial and descriptive frame elements from ophthalmology notes.
This QA-based method can be seen as a formulation of prompt-based models commonly seen in NLP \cite{taylor2022clinical, sivarajkumar2022healthprompt}.
In this method, we extract the spatial triggers and the main entities (e.g., eye finding, anatomical location) in the first turn and subsequently extract all the spatial (e.g., laterality) and descriptive (temporal descriptor or the timing of a finding) frame elements in the second turn.
Finally, we evaluate the performance of the two-turn QA system on a held-out test set of $100$ notes.
\textcolor{black}{We also investigate the potential benefit of transfer learning from a different medical domain (i.e., radiology) through sequential fine-tuning for our task of spatial information extraction.}

\section{Related work}
\label{sect:related_work}
We review relevant prior work in two categories.
Section \ref{sect:rel_work_datasets} focuses on the types of clinical text that previous research has explored for various ophthalmology NLP applications.
Section \ref{sect:rel_work_methods} presents an overview of the NLP methods proposed in the ophthalmology domain.

\subsection{Datasets and Applications in Ophthalmology NLP}
\label{sect:rel_work_datasets}
Wang et al. \cite{wang2022DeepLearningApproaches} used the first $3$ clinical progress notes from within the first $120$ days of follow-up along with structured clinical data for predicting whether glaucoma patients would require surgery.
Another study by Wang et al. \cite{wang2021LookingLowVision} used both structured and progress notes to predict visual prognosis.
Wang et al.\cite{wang2021DevelopmentEvaluationNovel} also demonstrated the usage of ophthalmology domain-specific word embeddings trained using PubMed abstracts and electronic health record notes to improve the performance of deep learning models in predicting visual prognosis when compared to using general word embeddings.
Baxter et al. \cite{baxter2020TextProcessingDetection} identified fungal ocular involvement cases using $26830$ free-text notes in the MIMIC-III clinical database.
Liu et al. \cite{liu2017NaturalLanguageProcessing} applied NLP on $743838$ operative notes to identify two key variables, intracameral antibiotic injection and posterior capsular rupture.
Stein et al. \cite{stein2019EvaluationAlgorithmIdentifying} used NLP to identify exfoliation syndrome from clinical notes.
Zheng et al. \cite{zheng2019UsingNaturalLanguage} applied an NLP algorithm on over 1 million clinical notes to identify herpes zoster ophthalmicus cases.

Mbagwu et al. \cite{mbagwu2016CreationAccurateAlgorithm} built a structured query language-based algorithm to first extract the Snellen visual acuities from structured laterality fields from $295218$ ophthalmology clinical notes and then capture the best documented visual acuity of each eye and this was evaluated against a clinician chart review of $100$ random notes.
Later, Baughman et al. \cite{baughman2017ValidationTotalVisual} developed a rule-based NLP algorithm to extract Snellen visual acuity from free-text inpatient ophthalmology notes collected from the University of Washington healthcare system in Seattle over an 8-year period and their algorithm was evaluated against the data points generated from manual review of $644$ notes.
Wang et al. \cite{wang2020AutomatedExtractionOphthalmica} developed rule-based algorithms to extract surgery outcome mentions representing implant usage--intraocular lens power and glaucoma implant type as well as surgery laterality from ophthalmology operative notes.
The algorithms were validated against manually-annotated random sets of $100$ operative notes for each of the three surgical categories and $100$ notes for laterality.
Maganti et al. \cite{maganti2019NaturalLanguageProcessing} designed an NLP algorithm to extract two quantitative key features of microbial keratitis--epithelial defect and stromal infiltrate as millimeter measurements from progress notes.
Recently, another work by Woodward et al. \cite{woodward2021DevelopmentValidationNatural} developed a rule-based NLP system to identify the clinical features of microbial keratitis, namely centrality, depth, and thinning  using the free text in the corneal examination section from physician notes.

We see that majority of the work in ophthalmology NLP is focused toward identifying cases associated with specific ocular diseases such as glaucoma and fungal ocular involvement.
Another strand of work attempted to extract certain information from the unstructured notes specifically related to visual acuity, surgery outcomes, and microbial keratitis.
From a dataset perspective, previous works mostly used operative, clinical, and progress notes while the two studies \cite{mbagwu2016CreationAccurateAlgorithm,baughman2017ValidationTotalVisual} on visual acuity extraction have utilized ophthalmology notes.
Thus, we find that limited research has focused on comprehensive information extraction across eye diseases and, therefore, in this work we attempt to extract more detailed clinical information from ophthalmology notes that can broaden the scope of applications in ophthalmology.

\subsection{NLP Methods for Ophthalmology Information Extraction}
\label{sect:rel_work_methods}
Most of the systems extracting ophthalmic information from unstructured notes developed rule-based methods.
Among these, Wang et al. \cite{wang2020AutomatedExtractionOphthalmica} used regular expressions to extract surgery outcomes.
Woodward et al. \cite{woodward2021DevelopmentValidationNatural} employed regular expressions, part-of-speech tagging and syntactic dependency parsing to extract features of microbial keratitis.
Mbagwu et al. \cite{mbagwu2016CreationAccurateAlgorithm} and Baughman et al. \cite{baughman2017ValidationTotalVisual} also developed rule-based algorithms for visual acuity extraction.
The former was based on structured query language where keyword search was performed on the structured laterality fields while the latter used regular expression in combination with additional rules.

Among the studies that focused on identifying certain ocular cases, Baxter et al. \cite{baxter2020TextProcessingDetection} developed a string matching method using regular expressions to extract text strings relevant to fungal ocular involvement.
Liu et al. \cite{liu2017NaturalLanguageProcessing} built a lexicon using SAS text-processing modules that identified misspellings, negations, and abbreviations and associated words to concepts for identifying the two key variables from operative notes.
Stein et al. \cite{stein2019EvaluationAlgorithmIdentifying} curated a list of terms and abbreviations to search for exfoliation syndrome-related mentions.
The search algorithm included identifying negated mentions based on surrounding text and used regular expressions and generalized Levenshtein edit distance to recognize misspellings.
Zheng et al. \cite{zheng2019UsingNaturalLanguage} created terminologies to search for herpes zoster-related information and also included relation detection algorithm for identifying herpes zoster ophthalmicus signs or symptoms.
A few studies \cite{wang2022DeepLearningApproaches,wang2021LookingLowVision,wang2021DevelopmentEvaluationNovel} by Wang et al. developed deep learning models for predicting glaucoma progression and whether low vision patients would have low vision after one year.
The models were based on convolutional neural network architectures and they used the previously developed ophthalmology domain-specific word embeddings \cite{wang2021DevelopmentEvaluationNovel} to represent the words.

Thus, we see that most of the existing methods are rule-based and they are restricted to specific entity extraction from ophthalmology-related text.
Only a few studies used deep learning-based methods and those too for prediction task.
In this work, we consider a wide variety of ophthalmic entities of clinical importance including findings, their locations,
and visual acuity scores.
Additionally, we also cover detailed spatial relations including relations between eye findings and locations as well as findings and their corresponding descriptors.
We employ an advanced transformer language-based question answering method to automatically extract the entities and relations.

\section{Data}

We use a set of $600$ notes for annotating the important ophthalmic entities and spatial relations.
These notes are collected from the Robert Cizik Eye Clinic at McGovern Medical School at Houston.
The notes contain information about a patient's history, detailed description of patients' experiences with their vision, interpretations of eye imaging examinations, information about past surgeries and their outcomes and complications, and associated neurological symptoms.
We use the BRAT tool \cite{stenetorp2012BratWebbasedTool} for annotation.

\subsection{Representation Schema}
\label{lab:annotation}
\begin{figure*}[!t]
\includegraphics[width=0.9\textwidth]{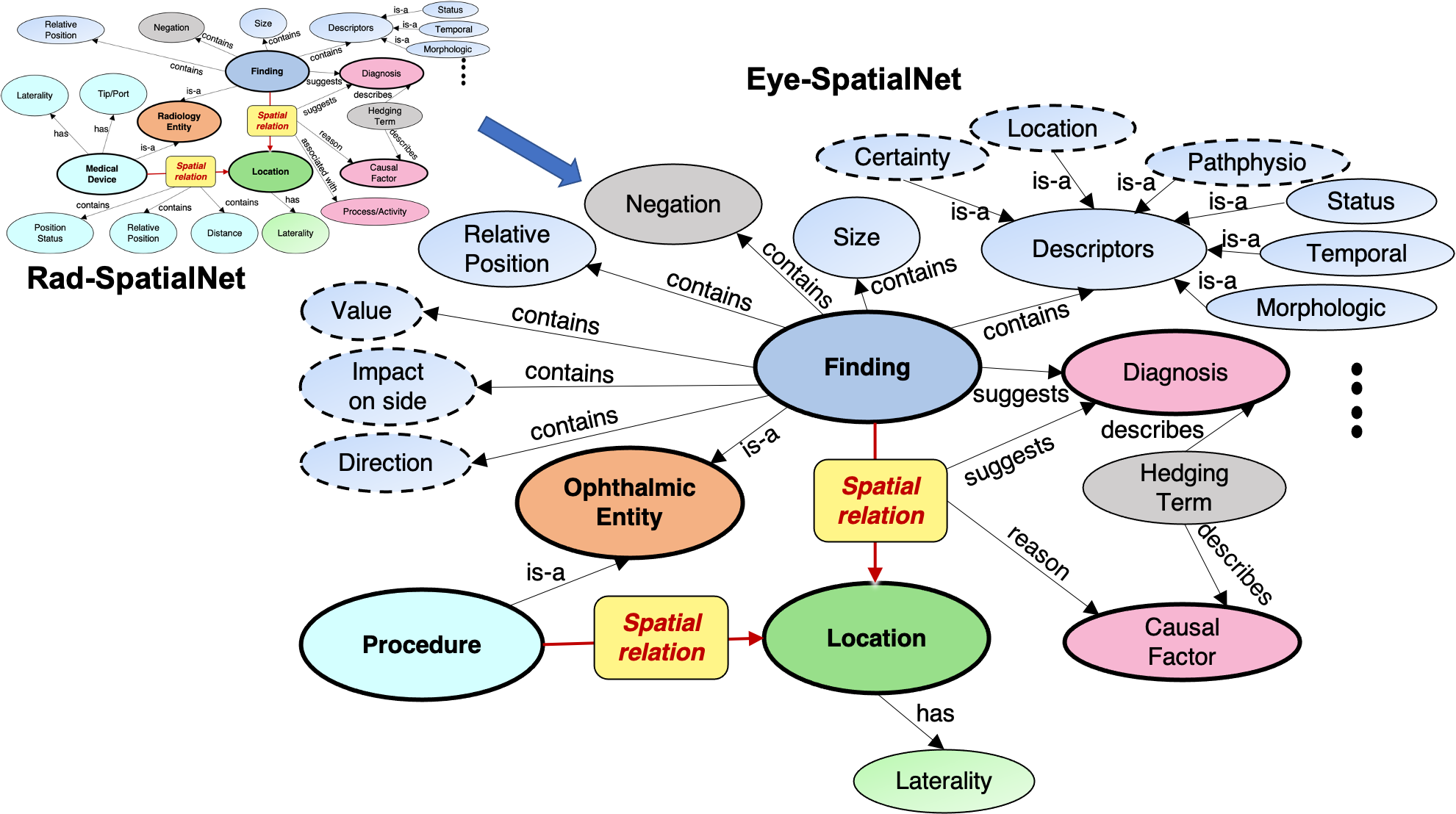}
\centering
\caption{Eye-SpatialNet schema. The dashed circles indicate the newly added frame elements.}
\label{fig:schema}
\end{figure*}

Our annotation schema is largely adopted from an existing frame-based spatial representation schema -- Rad-SpatialNet \cite{datta2020RadSpatialNetFramebasedResourceb}.
The spatial language encoded in the ophthalmology notes are different from those in radiology reports.
We represent the information in a way that can accurately capture ophthalmology-specific spatial meanings from the note text.
For this schema, we incorporate specific spatial and descriptive frame elements or relations besides the common ones proposed in Rad-SpatialNet.
The entity types included are spatial trigger, finding, anatomy,
device, location descriptor, other descriptor, assertion, quantity, drug, and procedure.
The spatial and descriptive frame elements are mostly similar to the ones described in our previous work \cite{datta2022FinegrainedSpatialInformationa}.
Additionally, we include the following frame elements: medication, impact on side, pathophysiologic descriptor, direction, associated diagnosis, specific location descriptor, certainty descriptor, and value.
Frame elements are either connected to the spatial trigger terms or the main clinical entities such as findings and anatomies.
We describe the newly added ophthalmology-specific frame elements in the following subsections.
The schema is illustrated in Figure \ref{fig:schema}.

\subsubsection{New spatial frame elements}
We add three new spatial frame elements related to findings, namely, exact location descriptor, impact on side, and direction.
For the exact location descriptor, let us consider the example below.
\begin{itemize}
    \item [] \textit{She was found to have 20/25 vision OD and CF vision OS with mild \textbf{disc} edema \textbf{in} the left eye.}
\end{itemize}
Here we see that there is a spatial relation between \textit{mild disc edema} and \textit{left eye} connected through the spatial trigger \textit{in}.
As per the Eye-SpatialNet schema, \textit{edema} has the spatial role of a `Figure' and its corresponding location \textit{eye} acts as the `Ground'.
Moreover, we notice that \textit{edema} has been described through a location descriptor \textit{disc} besides the status descriptor \textit{mild}.

Sometimes, a finding that has been detected in both sides (left and right) is described with different severity based on laterality or side.
\begin{itemize}
    \item [] \textit{External examination reveals a right relative proptosis with bilateral lid retraction \textbf{right greater than left}.}
\end{itemize}
In this example, \textit{retraction} is the finding that is more pronounced in the right eyelid than the left eyelid.
Moreover, \textit{retraction} is described using laterality \textit{bilateral} and location descriptor \textit{lid}.

A finding's direction is also documented in the notes.
\begin{itemize}
    \item [] \textit{She reports that her right eye deviated \textbf{outward}, and she had difficulty walking with poor coordination.}
\end{itemize}
Here, \textit{outward} is used to describe the direction of right eye deviation.

All these three frame elements--location descriptor, impact on side, and direction are associated with describing the detailed spatial aspects of a finding and, therefore, we include these elements in our representation schema.

\subsubsection{New descriptive frame elements}
Ophthalmologists often document detailed contextual information while describing the findings.
We add four descriptive frame elements related to findings, namely, certainty descriptor, associated diagnosis, pathophysiologic descriptor, and value.
Consider the example below.
\begin{itemize}
    \item [] \textit{His past ocular history is \textbf{significant} for optic neuritis and right optic atrophy.}
\end{itemize}
In this sentence, the term \textit{significant} is used to describe the certainty of both \textit{optic neuritis} and \textit{optic atrophy} findings.

Oftentimes, some findings are described along with their associated diagnoses.
In the following example, \textit{occlusions} is linked to \textit{Susac Syndrome}.
\begin{itemize}
    \item [] \textit{At this time the exact cause is unknown, however, with multiple retinal branch artery occlusions bilaterally one must entertain the diagnosis of \textbf{Susac Syndrome}.}
\end{itemize}
Note that this `Associated Diagnosis' frame element is different from the `Diagnosis' frame element proposed in Rad-SpatialNet \cite{datta2020RadSpatialNetFramebasedResourceb}.
The `Diagnosis' element is linked to a spatial trigger, whereas `Associated Diagnosis' element is linked to an eye finding (e.g., \textit{Susac Syndrome} in the sentence above).
In ``\textit{She did show me video of her episodes of upturning of the eyes which appears consistent with \textbf{oculogyric crisis}.}'', \textit{oculogyric crisis} acts as the `Diagnosis' element of the spatial frame instantiated by the spatial trigger \textit{of} connecting \textit{upturning} and \textit{eyes}.

We also include pathophysiologic descriptor of a finding in the schema.
For example, in ``\textit{She is seen in follow up for her left sided headache and retroorbital pain in the setting of presumed \textbf{autoimmune} retinopathy.}'', \textit{autoimmune} is the pathophysiologic descriptor associated with the finding \textit{retinopathy}.

Ophthalmology notes also contain information about visual acuity scores and other eye-related measurements.
We present two examples below.
\begin{enumerate}
    \item \textit{On his examination he found her to have \textbf{20/20} vision OD and \textbf{20/30} vision OS with a left RAPD.}
    \item \textit{INTRAOCULAR MEASUREMENT: Method: Applanation  Right Eye: \textbf{15}}
    \item \textit{LEFT EYE:  Media: hazy view  Cup/Disc Ratio: \textbf{0.4}}
\end{enumerate}
Note that in the first example, the first \textit{vision} occurrence has a visual acuity score of \textit{20/20} in the oculus dextrus (\textit{OD}) or the right eye, while the second vision has a score of \textit{20/30} in the oculus sinister (\textit{OS}) or the left eye.
Thus, the first \textit{vision} finding is linked to \textit{20/20} and the second \textit{vision} is linked to \textit{20/20} through the `Value' relation or frame element.
And \textit{OD} is the laterality of the first \textit{vision} while \textit{OS} is the laterality of the second one.
In the second example, \textit{INTRAOCULAR MEASUREMENT} is linked to \textit{15} through the `Value' element, whereas the third example shows that the finding \textit{Cup/Disc Ratio} is associated with its corresponding value \textit{0.4}.
Therefore, we capture all the important eye measurements in our schema.

Apart from above additions, this schema covers temporal information of findings that are expressed using a variety of phrases unlike the temporal descriptors of radiological findings annotated in Rad-SpatialNet \cite{datta2020RadSpatialNetFramebasedResourceb}.
These expressions include \textit{one and a half to two years}, \textit{$>$ 8 years}, \textit{next 3-4 months}, \textit{within 1-2 months post-operatively}, \textit{over the next few days}, and \textit{early in the mornings}.
This schema covers lateralities that are specific to ophthalmology such as \textit{OS}, \textit{OD}, and \textit{OU}, besides the common ones such as \textit{left}, \textit{right}, and \textit{bilateral}.

\begin{table}[t]
    \caption{Basic statistics. Avg - Average.} \label{tab:basic_stats}
	\centering
	\begin{tabular}{l c}
		\hline 
	  \textbf{Item} & \textbf{Value} \\  
	  \hline
	  Avg. note length (in tokens) & $470.61$ \\
      Avg. sentence length (in tokens) & $20.34$ \\
      No. of unique spatial triggers & $49$ \\
		\hline
	\end{tabular}
 \end{table}

 \begin{table}[t]
\caption{Main entities.} \label{tab:entities}
	\centering
	\begin{tabular}{l c c}
		\hline 
	  \textbf{Entity type} & \textbf{Frequency} & \textbf{F1 agreement} \\  
	  \hline
	  Spatial trigger & $1715$ & $0.91$ \\
      Finding & $7308$ & $0.80$  \\
      Anatomy & $2424$ & $0.88$  \\
      Device & $14$ & $0.90$ \\
      Drug & $22$ & $0.60$  \\
      Procedure & $182$ & $0.35$ \\
      Other descriptor & $9782$ & $0.79$ \\
      Quantity & $366$ & $0.88$ \\
      Assertion & $1616$ & $0.70$ \\
      Location descriptor & $132$ & $0.60$ \\ 
		\hline
	\end{tabular}
\end{table}

\begin{table}[t]
\caption{Spatial frame elements. 
	} \label{tab:spatial_FEs}
	\centering
	\begin{tabular}{l c c}
	\hline 
	  \textbf{Frame element} & \textbf{Frequency} & \textbf{F1 agreement} \\  
	  \hline
	  Figure & $2261$ & $0.77$ \\
      Ground & $2094$ & $0.89$  \\
      Hedge & $397$ & $0.69$  \\
      Diagnosis & $18$ & $0.28$ \\
      Relative Position & $132$ & $0.59$  \\
      Reason & $7$ & $0.77$ \\
      Medication & $18$ & $0.64$ \\
      Morphologic & $45$ & $0.44$ \\
      Size Desc & $43$ & $0.56$ \\
      Distribution Pattern & $83$ & $0.29$ \\
      Composition & $36$ & $0.59$ \\
      Laterality & $3464$ & $0.78$ \\
      Size & $48$ & $0.30$ \\
      Impact on Side & $97$ & $0.75$ \\
      Direction & $85$ & $0.56$ \\
      Specific location & $1636$ & $0.72$ \\
		\hline
	\end{tabular}
\end{table}

\begin{table}[t]
	\caption{Descriptive frame elements.
	} \label{tab:descriptive_FEs}
	\centering
		\begin{tabular}{l c c}
		\hline 
	  \textbf{Frame element} & \textbf{Frequency} & \textbf{F1 agreement} \\  
	  \hline
	  Status & $3051$ & $0.59$ \\
      Quantity & $101$ & $0.56$  \\
      Temporal & $1066$ & $0.45$  \\
      Negation & $921$ & $0.55$ \\
      Pathphysio & $75$ & $0.60$  \\
      Certainty & $298$ & $0.49$ \\
      Associated Diagnosis & $72$ & $0.23$ \\
      Value & $318$ & $0.83$ \\
		\hline
	\end{tabular}
\end{table}

\subsection{Annotation statistics}
Each note was annotated by two annotators having medical background (one optometrist, one MD) and the annotations were reconciled iteratively through discussions.
The overall F1 agreements are reported for annotating the main entities, the spatial and descriptive frame elements.
We show the statistics of our annotated dataset as well as the inter-annotator agreement measures in Tables \ref{tab:basic_stats}, \ref{tab:entities}, \ref{tab:spatial_FEs}, and \ref{tab:descriptive_FEs}.
The average sentence length ($20.34$) of the ophthalmology notes is slightly higher than that of the radiology reports dataset \cite{datta2020RadSpatialNetFramebasedResourceb}.
The spatial triggers in our annotated ophthalmology dataset contains spatial prepositions such as \textit{in}, \textit{behind}, and \textit{within} as well as verbs such as \textit{appear}, \textit{reveals}, and \textit{are}.
The top three frequent trigger terms are \textit{in}, \textit{of}, and \textit{are}.
Among the entity types, the agreement is low (F1: $0.35$) for Procedure as this involves identifying different eye surgery procedures or therapies that are often expressed in their abbreviated forms (e.g., LPI for Laser Peripheral Iridotomy and PRP for Pan-retinal photocoagulation).
Among the spatial and descriptive frame elements, Diagnosis, Size, and Associated Diagnosis have low F1 agreements of $0.28$, $0.30$, and $0.23$, respectively, as it is oftentimes difficult to correctly interpret the potential diagnosis terms and sizes of different eye entities.
Diagnosis terms are difficult to differentiate from the Finding terms and are often annotated as Findings.
Another general challenge in the annotation process involved separating the eye-related findings from the neuroradiological findings as oftentimes the interpretations of brain images are embedded in the ophthalmology notes.

\section{Methods}
\subsection{Overview}
We frame the task of spatial information extraction (IE) from ophthalmology notes as two-turn question answering (QA).
This formulation (both single and multi turn QA) has proven to perform well for various general and biomedical domain IE tasks.
Our previous work has also demonstrated the improved performance of a two-turn QA framework over a more standard sequence labeling-based method to extract detailed information from radiology text \cite{datta2022FinegrainedSpatialInformationa}.
Inspired by these findings, we adopt a similar two-turn QA approach to identify the spatial triggers, the main ophthalmic entities, and their corresponding spatial and descriptive frame elements.
This framework is suitable for IE scenarios where identification of relations or frame elements are dependent on extracting the target entities or lexical units of the frames (i.e., spatial triggers and ophthalmic entities).
In this, the aim is to query a machine reading comprehension (MRC) model for returning answers given a query and the context passage (ophthalmology note text).
The MRC system is based on the pre-trained language model BERT \cite{devlin2019bert} where we follow the standard BERT input format by combining the query and the note text.
The system allows for multiple answer extraction against a query, which is suitable for our schema as there can be multiple frame elements of the same type that are linked to a particular entity (spatial trigger or other ophthalmic entity).
The MRC framework involving two BERT models for the two turns are adopted from a previous work \cite{li2019EntityRelationExtractionMultiTurna}.

\subsection{Query generation}
We construct queries for the newly added entities and frame elements in Eye-SpatialNet.
We adopt the same query templates for both target entity and element extraction as used in our previous work \cite{datta2022FinegrainedSpatialInformationa}.
Queries for the first turn incorporate the entity types whereas queries for the second turn include information about the frame elements and the associated main entity that is extracted in the first turn.
In this paper, we use the Query\textsubscript{find + desc} variant to extract the frame elements in the second turn.
The idea is to make the query more informative through incorporation of domain knowledge by adding a description of the particular frame element of interest at the beginning of a query.
The following is an example query to extract `ImpactOnSide' spatial element.
\begin{itemize}
    \item [] \textit{ImpactOnSide refers to which eye side is more impacted. Examples include right greater than left, smaller than left, and worse in the left eye. find all \textbf{descriptor} entities in the context that have a \textbf{impact on side} relationship with \textbf{clinical finding} entity \ul{optic neuropathy}.}
\end{itemize}
Here, we see that the query includes description about ImpactOnSide as well as the finding entity (i.e., \textit{optic neuropathy}) that is identified in the previous turn.
If no answer is retrieved from the MRC system, this means there is no such entity of type `Descriptor' in the note text that captures information about which eye side is more or less affected by \textit{optic neuropathy}.
The descriptions used to form the queries for all new frame elements are shown in Table \ref{table:query_descriptions}.

\begin{table}[t]
\caption{Descriptions used in the queries to extract additional frame elements.} 
\centering
\begin{tabular}{p{0.2\linewidth} p{0.7\linewidth}}
      \hline
      \textbf{Frame element} & \textbf{Description} \\
      \hline
      Medication & Medication refers to a drug or solution that has been administered or applied to any eye location. \\
      \hline
      ImpactOnSide & ImpactOnSide refers to which eye side is more impacted. Examples include right greater than left, smaller than left, and worse in the left eye. \\
      \hline
      PathphysioDesc & Pathophysiologic descriptor refers to the functional changes that accompany a disease. Examples include autoimmune and physiologic. \\
      \hline
      Direction & Direction indicates direction of a finding. Examples include outward and to the right.\\
      \hline
      AssocDiag & Associated diagnosis refers to the clinical condition or disease associated with a finding. This usually appears after phrases such as associated with and secondary to.\\
      \hline
      LocationDesc & Location descriptor refers to the exact location of a finding. Examples include retrooorbital and optic disc. \\
      \hline
      CertaintyDesc & Certainty descriptor refers to uncertainty phrases describing a finding. Examples include significant and consistent with. \\
      \hline
      Value & Value refers to a visual acuity score or any measurement or ratio. Examples include 20/20, 20/40, 16, and 0.8.\\
      \hline
\end{tabular}
\label{table:query_descriptions}
\end{table}

\begin{table}[t]
    \caption{Target entity extraction results using BERT\textsubscript{LARGE}-MIMIC two-turn QA method. desc - Descriptor.}
	\centering
\begin{tabular}{lccc}
      \hline
      \textbf{\textsc{Entity}} & \textbf{Precision(\%)} & \textbf{Recall (\%)} & \textbf{F1} \\
      \hline
      Spatial trigger & $86.86$ & $91.89$ &  $89.31$ \\
      Finding & $75.71$ & $83.41$ & $79.37$ \\
      Anatomy & $85.37$ & $85.15$ & $85.26$ \\
      Location desc & $30.77$ & $40.00$ & $34.78$ \\
      Other desc & $76.57$ & $83.04$ & $79.67$ \\
      Assertion & $81.78$ & $89.80$ & $85.60$ \\
      Quantity & $82.89$ & $82.89$ & $82.89$ \\
      Procedure & $56.67$ & $53.12$ & $54.84$ \\
      \hline
\end{tabular}
\label{table:bert_ent_extraction}
\end{table}

\begin{table}[t]
    \caption{Frame element extraction results using BERT\textsubscript{LARGE}-MIMIC two-turn QA method. sptr - Spatial trigger. Desc - Descriptive.} \label{table:bert_fe_extraction}
      \centering
      \begin{tabular}{c|lccc}      
      \hline
      \multicolumn{2}{c}{\textbf{\textsc{Frame Elements}}} & \textbf{Precision(\%)} & \textbf{Recall (\%)} & \textbf{F1} \\
      \hline
      \multirow{6}{*}{\rotatebox{90}{\parbox{2cm}{\centering \small\textbf{Spatial(sptr)}}}} & Figure & $75.29$ & $74.43$ &  $74.86$ \\
      & Ground & $85.89$ & $91.21$ & $88.47$ \\
      & Hedge & $89.47$ & $86.44$ & $87.93$ \\
      & Relative Position & $30.43$ & $70.00$ & $42.42$ \\
      & Medication & $50.00$ & $100$ & $66.67$ \\
      \hline
      \multirow{7}{*}{\rotatebox{90}{\parbox{3cm}{\centering \small\textbf{Spatial(entity)}}}} & Laterality & $80.59$ & $83.15$ &  $81.85$ \\
      & Distribution Pattern & $47.37$ & $64.29$ & $54.55$ \\
      & SizeDesc & $60.00$ & $42.86$ & $50.00$ \\
      & LocationDesc & $69.26$ & $76.21$ & $72.57$ \\
      & ImpactOnSide & $72.73$ & $84.21$ & $78.05$ \\
      & Direction & $57.14$ & $66.67$ & $61.54$ \\
      & Size & $28.57$ & $10.00$ & $14.81$ \\
      \hline
      \multirow{7}{*}{\rotatebox{90}{\parbox{3cm}{\centering \small\textbf{Desc(entity)}}}} & Status & $70.11$ & $70.93$ &  $70.52$ \\
      & Quantity & $63.64$ & $43.75$ & $51.85$ \\
      & Temporal & $53.33$ & $43.78$ & $48.09$ \\
      & Negation & $77.60$ & $82.32$ & $79.89$ \\
      & Certainty & $60.26$ & $64.38$ & $62.25$ \\
      & Pathphysio & $47.06$ & $53.33$ & $50.00$ \\
      & Value & $81.63$ & $68.97$ & $74.77$ \\
      \hline
\end{tabular}
\end{table}

\begin{table}[t]
    \caption{F1 measures for different fine-tuning variations using BERT\textsubscript{LARGE}-MIMIC sequence labeling method on $100$ test ophthalmology notes. \textbf{Eye}: Fine-tuning only on Eye-SpatialNet (Ophthalmology), \textbf{Rad$\rightarrow$Eye}: Fine-tuning on Rad-SpatialNet (Radiology) followed by Eye-SpatialNet, \textbf{Rad}: Fine-tuning only on Rad-SpatialNet.
      } \label{table:seq_fine_tuning}
      \centering
      \begin{tabular}{lccc}
      \hline
      \textbf{\textsc{Frame Element}} & \textbf{Eye} & \textbf{Rad$\rightarrow$Eye} & \textbf{Rad} \\
      \hline
      Figure & $78.76$ & $80.88$ & $51.29$ \\
      Ground & $95.38$ & $95.19$ & $91.95$ \\
      Hedge & $82.64$ & $89.08$ & $0$ \\
      Relative Position & $60.87$ & $57.14$ & $43.48$ \\
      \hline
\end{tabular}
\end{table}

\section{Experimental Settings and Evaluation}
We randomly split our annotated ophthalmology dataset of $600$ notes such that $450$ notes are used for training, $50$ for development, and $100$ for testing.
We use a clinical BERT\textsubscript{LARGE} model that is pre-trained on MIMIC-III clinical notes for 300K steps \cite{si2019EnhancingClinicalConcept} 
as it performed better on the radiology reports dataset \cite{datta2022FinegrainedSpatialInformationa}.
We fine-tune BERT\textsubscript{LARGE}-MIMIC (cased version) on our Eye-SpatialNet dataset for $10$ epochs and use the same hyperparameter settings as reported in Datta et al. \cite{datta2022FinegrainedSpatialInformationa}.
We evaluate the performance metrics - precision, recall, and F1 score and report the results on the test set of $100$ notes.
We consider exact matches of the entity and frame element spans against the annotated spans for evaluation.

Further, to leverage an already available language model that is fine-tuned on the radiology reports dataset (introduced in Datta et al. \cite{datta2020RadSpatialNetFramebasedResourceb}) for the task of spatial information extraction, we evaluate any prospective benefits of transfer learning through sequential fine-tuning, that is, by first fine-tuning the model on radiology reports followed by fine-tuning on ophthalmology notes.
The radiology fine-tuning was performed on $288$ reports and we further fine-tune on $450$ ophthalmology notes.
For this, we use the BERT\textsubscript{LARGE}-MIMIC sequence labeling model from Datta et al. \cite{datta2020RadSpatialNetFramebasedResourceb}.
Note that we use the gold spatial triggers for this experiment to extract the elements that are connected to the triggers (and not the main ophthalmic entities).
Using predicted triggers would provide a more realistic evaluation, but that is not the focus of this experiment.
We evaluate the performance on the main spatial frame elements that are common between the two domains on the $100$ test ophthalmology notes.
For fine-tuning the sequence labeling model on the ophthalmology data, we set the maximum sequence length at $128$, learning rate at $2e-5$, and number of training epochs at $4$.

\section{Results}
The performance measures for extracting the main ophthalmic entities in the first turn from $100$ ophthalmology test notes are reported in Table \ref{table:bert_ent_extraction}.
The results are promising for the common entities including `Spatial trigger', `Finding', and `Anatomy' with F1 scores of $89.31$, $79.37$, and $85.26$, respectively, while they are low for `Location descriptor' and `Procedure'.
Note that the entities `Drug' and `Device' occur very infrequently in the dataset (with only $2$ and $1$ occurrences in the test set) and the performance measures are zero.

We show the results for extracting the spatial and descriptive frame elements in the second turn in Table \ref{table:bert_fe_extraction}.
We see that the model performs well for common frame elements such as `Ground', `Hedge', `Laterality', `ImpactOnSide', and `Negation'.
The performance measures are particularly low for `Relative Position', `Size', and `Temporal Desc'.
This may be because of the wide variation in the phrases used to express the sizes and temporalities of findings.
Moreover, there are only $3$, $1$, and $7$ instances of the frame elements `Diagnosis', `Reason', and `Composition Desc', respectively in the test set with no occurrence for `Morphologic Desc'.
Although the element `AssocDiag' occurs $21$ times, the performance values are zero for this element.
The reason could be that this is often a difficult task (even for humans) to differentiate these terms from findings.

The results of transfer learning experiment from radiology to ophthalmology domain is shown in Table \ref{table:seq_fine_tuning}.
We see the F1 scores improve from $78.76$ to $80.88$ for `Figure' and from $82.64$ to $89.08$ for `Hedge' when we use a model fine-tuned on radiology reports to further fine-tune on our ophthalmology dataset.
We also note that the F1 measure for the `Ground' element is $91.95$ without the requirement of any fine-tuning on ophthalmology data.
The results are zero for `Diagosis' and `Reason' as they are too infrequent in the dataset as stated above.

\section{Discussion}
We present a new dataset of $600$ ophthalmology notes annotated with detailed spatial and contextual information.
Although a few studies worked on identifying a certain set of entities from clinical notes, they are mostly focused toward visual acuity and features of microbial keratitis \cite{baughman2017ValidationTotalVisual,mbagwu2016CreationAccurateAlgorithm,woodward2021DevelopmentValidationNatural}.
Our work is an initial effort in building a schema that captures more detailed information from the notes that can potentially be used in various useful ophthalmology-related applications and research studies.

Most of the entities and frame elements used in encoding spatial language in the ophthalmology notes are adopted from our previously proposed Rad-SpatialNet schema \cite{datta2020RadSpatialNetFramebasedResourceb} built for radiology.
This indicates the generalizability of the schema in that it captures most of the common and important spatial information usually encountered in clinical text.
In this work, we incorporate additional frame elements for two reasons. First, to cover more detailed information about the findings that were not present in Rad-SpatialNet such as capturing implicit spatial relations through the `Location Desc' frame element (e.g., scenarios where a spatial relation exists but a spatial trigger term is not present in the sentence).
Second, to include ophthalmology-specific spatially-grounded entities (e.g., `Procedure') and elements that are of interest to ophthalmology researchers (e.g., visual acuity and other important eye measurements through the `Value' frame element).
The results in Tables \ref{table:bert_ent_extraction} and \ref{table:bert_fe_extraction} show that the two-turn QA approach achieves satisfactory performance in identifying different entities and frame elements and are comparable to the results on the radiology report dataset \cite{datta2022FinegrainedSpatialInformationa}.
We achieve this without any modification of the query templates and the frame element descriptions (that are used to form the queries) for those elements that also exist in Rad-SpatialNet.
This also indicates that the method is adaptable and generalizable enough to work satisfactorily well for frequent entity types and frame elements across medical domains (although the language style and the vocabulary differ substantially between radiology reports and ophthalmology notes).

To examine the effect of transfer learning from a different medical domain, our experiment with the sequence labeling model in Table \ref{table:seq_fine_tuning} indicates that transfer learning holds potential in improving the performance for some frame elements, however, a more thorough evaluation covering all other elements is required to understand its real benefits.
This includes experimenting with a small number of ophthalmology notes in the fine-tuning process, as often only a limited amount of labeled data is available in a new domain.
Interestingly, although the two-turn QA approach works well both for ophthalmology and radiology domains, our initial experiments with sequential fine-tuning did not yield good results using the QA approach.
We leave this to future work where we plan to investigate this further and evaluate the less explored domain adaptation techniques such as the adaptive off-the-shelf approach proposed in Laparra et al. \cite{laparra2020RethinkingDomainAdaptation}.

To handle less frequent entities and frame elements better as well as to further improve the QA model's performance, we plan to augment the dataset by automatically generating a large weakly labeled ophthalmology dataset using domain-specific rules, a technique that has been validated to be useful by many recent studies in the medical domain \cite{smit2020CombiningAutomaticLabelersa,fries2021OntologydrivenWeakSupervisionb}.
Apart from reducing the annotation effort, this can particularly be useful for elements such as `Size' and `Value' that usually follow a set of patterns based on domain.
For example, `$4$--$>3$mm' is used to express pupil size in an ophthalmology note whereas `$2.1$ x $3.4$ x $2.0$ cm' denotes a tumor size in a radiology report.
Finally, for a more exhaustive evaluation on this proposed dataset, we also intend to incorporate cross validation in a later work.

\section{Conclusion}
We annotated $600$ ophthalmology notes with important spatial and contextual information of clinical importance.
We adopt our previously proposed Rad-SpatialNet schema and incorporate additional ophthalmology-specific information to encode spatial language in ophthalmology.
We apply a well-established approach of framing the extraction task as question answering to automatically identify the ophthalmic entities and their associated spatial and descriptive frame elements .
Our two-turn QA method performed well with high F1 scores for common entities and elements.

\section*{Funding}
This work was supported in part by the National Institute of Biomedical Imaging and Bioengineering (NIBIB: R21EB029575), the Patient-Centered Outcomes Research Institute (PCORI: ME-2018C1-10963) and the Cancer Prevention and Research Institute of Texas (CPRIT RP210045).

\bibliography{ophthalmology}

\begin{thebibliography}{10}
\expandafter\ifx\csname url\endcsname\relax
  \def\url#1{\texttt{#1}}\fi
\expandafter\ifx\csname urlprefix\endcsname\relax\def\urlprefix{URL }\fi
\expandafter\ifx\csname href\endcsname\relax
  \def\href#1#2{#2} \def\path#1{#1}\fi

\bibitem{wang2022DeepLearningApproaches}
S.~Wang, B.~Tseng, T.~{Hernandez-Boussard}, Deep {{Learning Approaches}} for
  {{Predicting Glaucoma Progression Using Electronic Health Records}} and
  {{Natural Language Processing}}, Ophthalmology Science 0~(0).
\newblock \href {http://dx.doi.org/10.1016/j.xops.2022.100127}
  {\path{doi:10.1016/j.xops.2022.100127}}.

\bibitem{zheng2019UsingNaturalLanguage}
C.~Zheng, Y.~Luo, C.~Mercado, L.~Sy, S.~J. Jacobsen, B.~Ackerson, B.~Lewin,
  H.~F. Tseng, Using natural language processing for identification of herpes
  zoster ophthalmicus cases to support population-based study, Clinical \&
  Experimental Ophthalmology 47~(1) (2019) 7--14.
\newblock \href {http://dx.doi.org/10.1111/ceo.13340}
  {\path{doi:10.1111/ceo.13340}}.

\bibitem{stein2019EvaluationAlgorithmIdentifying}
J.~D. Stein, M.~Rahman, C.~Andrews, J.~R. Ehrlich, S.~Kamat, M.~Shah, E.~A.
  Boese, M.~A. Woodward, J.~Cowall, E.~H. Trager, P.~Narayanaswamy, D.~A.
  Hanauer, Evaluation of an {{Algorithm}} for {{Identifying Ocular Conditions}}
  in {{Electronic Health Record Data}}, JAMA ophthalmology 137~(5) (2019)
  491--497.
\newblock \href {http://dx.doi.org/10.1001/jamaophthalmol.2018.7051}
  {\path{doi:10.1001/jamaophthalmol.2018.7051}}.

\bibitem{baughman2017ValidationTotalVisual}
D.~M. Baughman, G.~L. Su, I.~Tsui, C.~S. Lee, A.~Y. Lee, Validation of the
  {{Total Visual Acuity Extraction Algorithm}} ({{TOVA}}) for {{Automated
  Extraction}} of {{Visual Acuity Data From Free Text}}, {{Unstructured
  Clinical Records}}, Translational Vision Science \& Technology 6~(2) (2017)
  2.
\newblock \href {http://dx.doi.org/10.1167/tvst.6.2.2}
  {\path{doi:10.1167/tvst.6.2.2}}.

\bibitem{mbagwu2016CreationAccurateAlgorithm}
M.~Mbagwu, D.~D. French, M.~Gill, C.~Mitchell, K.~Jackson, A.~Kho, P.~J. Bryar,
  Creation of an {{Accurate Algorithm}} to {{Detect Snellen Best Documented
  Visual Acuity}} from {{Ophthalmology Electronic Health Record Notes}}, JMIR
  Medical Informatics 4~(2) (2016) e14.
\newblock \href {http://dx.doi.org/10.2196/medinform.4732}
  {\path{doi:10.2196/medinform.4732}}.

\bibitem{woodward2021DevelopmentValidationNatural}
M.~A. Woodward, N.~Maganti, L.~M. Niziol, S.~Amin, A.~Hou, K.~Singh,
  Development and {{Validation}} of a {{Natural Language Processing Algorithm}}
  to {{Extract Descriptors}} of {{Microbial Keratitis From}} the {{Electronic
  Health Record}}, Cornea 40~(12) (2021) 1548--1553.
\newblock \href {http://dx.doi.org/10.1097/ICO.0000000000002755}
  {\path{doi:10.1097/ICO.0000000000002755}}.

\bibitem{datta2020RadSpatialNetFramebasedResourceb}
S.~Datta, M.~Ulinski, J.~{Godfrey-Stovall}, S.~Khanpara, R.~F.
  {Riascos-Castaneda}, K.~Roberts, Rad-{{SpatialNet}}: {{A Frame-based
  Resource}} for {{Fine-Grained Spatial Relations}} in {{Radiology Reports}},
  in: Proceedings of the 12th {{Language Resources}} and {{Evaluation
  Conference}}, 2020, pp. 2251--2260.

\bibitem{li2019EntityRelationExtractionMultiTurna}
X.~Li, F.~Yin, Z.~Sun, X.~Li, A.~Yuan, D.~Chai, M.~Zhou, J.~Li,
  Entity-{{Relation Extraction}} as {{Multi-Turn Question Answering}}, in:
  Proceedings of the 57th {{Annual Meeting}} of the {{ACL}}, 2019, pp.
  1340--1350.
\newblock \href {http://dx.doi.org/10.18653/v1/P19-1129}
  {\path{doi:10.18653/v1/P19-1129}}.

\bibitem{datta2022FinegrainedSpatialInformationa}
S.~Datta, K.~Roberts, Fine-grained spatial information extraction in radiology
  as two-turn question answering, International Journal of Medical Informatics
  158 (2022) 104628.
\newblock \href {http://dx.doi.org/10.1016/j.ijmedinf.2021.104628}
  {\path{doi:10.1016/j.ijmedinf.2021.104628}}.

\bibitem{devlin2019bert}
J.~Devlin, M.-W. Chang, K.~Lee, K.~Toutanova, {{BERT}}: {{Pre-training}} of
  {{Deep Bidirectional Transformers}} for {{Language Understanding}}, in:
  Proceedings of the 2019 {{NAACL-HLT}}, 2019, pp. 4171--4186.
\newblock \href {http://dx.doi.org/10.18653/v1/N19-1423}
  {\path{doi:10.18653/v1/N19-1423}}.

\bibitem{li2020UnifiedMRCFramework}
X.~Li, J.~Feng, Y.~Meng, Q.~Han, F.~Wu, J.~Li, A {{Unified MRC Framework}} for
  {{Named Entity Recognition}}, in: Proceedings of the 58th {{Annual Meeting}}
  of the {{Association}} for {{Computational Linguistics}}, 2020, pp.
  5849--5859.
\newblock \href {http://dx.doi.org/10.18653/v1/2020.acl-main.519}
  {\path{doi:10.18653/v1/2020.acl-main.519}}.

\bibitem{levy2017ZeroShotRelationExtraction}
O.~Levy, M.~Seo, E.~Choi, L.~Zettlemoyer, Zero-{{Shot Relation Extraction}} via
  {{Reading Comprehension}}, in: Proceedings of the 21st {{Conference}} on
  {{Computational Natural Language Learning}} ({{CoNLL}} 2017), 2017, pp.
  333--342.
\newblock \href {http://dx.doi.org/10.18653/v1/K17-1034}
  {\path{doi:10.18653/v1/K17-1034}}.

\bibitem{liu2020EventExtractionMachine}
J.~Liu, Y.~Chen, K.~Liu, W.~Bi, X.~Liu, Event {{Extraction}} as {{Machine
  Reading Comprehension}}, in: Proceedings of the 2020 {{Conference}} on
  {{Empirical Methods}} in {{Natural Language Processing}} ({{EMNLP}}), 2020,
  pp. 1641--1651.
\newblock \href {http://dx.doi.org/10.18653/v1/2020.emnlp-main.128}
  {\path{doi:10.18653/v1/2020.emnlp-main.128}}.

\bibitem{li2020EventExtractionMultiturn}
F.~Li, W.~Peng, Y.~Chen, Q.~Wang, L.~Pan, Y.~Lyu, Y.~Zhu, Event {{Extraction}}
  as {{Multi-turn Question Answering}}, in: Findings of the {{Association}} for
  {{Computational Linguistics}}: {{EMNLP}} 2020, 2020, pp. 829--838.
\newblock \href {http://dx.doi.org/10.18653/v1/2020.findings-emnlp.73}
  {\path{doi:10.18653/v1/2020.findings-emnlp.73}}.

\bibitem{wang2020BiomedicalEventExtraction}
X.~D. Wang, L.~Weber, U.~Leser, Biomedical {{Event Extraction}} as {{Multi-turn
  Question Answering}}, in: Proceedings of the 11th {{International Workshop}}
  on {{Health Text Mining}} and {{Information Analysis}}, 2020, pp. 88--96.
\newblock \href {http://dx.doi.org/10.18653/v1/2020.louhi-1.10}
  {\path{doi:10.18653/v1/2020.louhi-1.10}}.

\bibitem{taylor2022clinical}
N.~Taylor, Y.~Zhang, D.~Joyce, A.~Nevado-Holgado, A.~Kormilitzin, Clinical
  prompt learning with frozen language models (2022).
\newblock \href {http://arxiv.org/abs/2205.05535} {\path{arXiv:2205.05535}}.

\bibitem{sivarajkumar2022healthprompt}
S.~Sivarajkumar, Y.~Wang, Healthprompt: A zero-shot learning paradigm for
  clinical natural language processing (2022).
\newblock \href {http://arxiv.org/abs/2203.05061} {\path{arXiv:2203.05061}}.

\bibitem{wang2021LookingLowVision}
S.~Y. Wang, B.~Tseng, Looking for {{Low Vision}}: {{Deep Learning}} and
  {{Natural Language Processing}} to {{Predict Visual Prognosis}},
  Investigative Ophthalmology \& Visual Science 62~(8) (2021) 3502.

\bibitem{wang2021DevelopmentEvaluationNovel}
S.~Wang, B.~Tseng, T.~{Hernandez-Boussard}, Development and evaluation of novel
  ophthalmology domain-specific neural word embeddings to predict visual
  prognosis, International Journal of Medical Informatics 150 (2021) 104464.
\newblock \href {http://dx.doi.org/10.1016/j.ijmedinf.2021.104464}
  {\path{doi:10.1016/j.ijmedinf.2021.104464}}.

\bibitem{baxter2020TextProcessingDetection}
S.~L. Baxter, A.~R. Klie, B.~R. Saseendrakumar, G.~Y. Ye, M.~Hogarth, Text
  {{Processing}} for {{Detection}} of {{Fungal Ocular Involvement}} in
  {{Critical Care Patients}}: {{Cross-Sectional Study}}, JMIR 22~(8) (2020)
  e18855.
\newblock \href {http://dx.doi.org/10.2196/18855} {\path{doi:10.2196/18855}}.

\bibitem{liu2017NaturalLanguageProcessing}
L.~Liu, N.~H. Shorstein, L.~B. Amsden, L.~J. Herrinton, Natural language
  processing to ascertain two key variables from operative reports in
  ophthalmology, Pharmacoepidemiology and Drug Safety 26~(4) (2017) 378--385.
\newblock \href {http://dx.doi.org/10.1002/pds.4149}
  {\path{doi:10.1002/pds.4149}}.

\bibitem{wang2020AutomatedExtractionOphthalmica}
S.~Y. Wang, S.~Pershing, E.~Tran, T.~{Hernandez-Boussard}, Automated extraction
  of ophthalmic surgery outcomes from the electronic health record,
  International Journal of Medical Informatics 133 (2020) 104007.
\newblock \href {http://dx.doi.org/10.1016/j.ijmedinf.2019.104007}
  {\path{doi:10.1016/j.ijmedinf.2019.104007}}.

\bibitem{maganti2019NaturalLanguageProcessing}
N.~Maganti, H.~Tan, L.~M. Niziol, S.~Amin, A.~Hou, K.~Singh, D.~Ballouz, M.~A.
  Woodward, Natural {{Language Processing}} to {{Quantify Microbial Keratitis
  Measurements}}, Ophthalmology 126~(12) (2019) 1722--1724.
\newblock \href {http://dx.doi.org/10.1016/j.ophtha.2019.06.003}
  {\path{doi:10.1016/j.ophtha.2019.06.003}}.

\bibitem{stenetorp2012BratWebbasedTool}
P.~Stenetorp, S.~Pyysalo, G.~Topi{\'c}, T.~Ohta, S.~Ananiadou, J.~Tsujii, Brat:
  A {{Web-based Tool}} for {{NLP-Assisted Text Annotation}}, in: Proceedings of
  the {{Demonstrations}} at the 13th {{EACL}}, 2012, pp. 102--107.

\bibitem{si2019EnhancingClinicalConcept}
Y.~Si, J.~Wang, H.~Xu, K.~Roberts, Enhancing clinical concept extraction with
  contextual embeddings, Journal of the American Medical Informatics
  Association 26~(11) (2019) 1297--1304.
\newblock \href {http://dx.doi.org/10.1093/jamia/ocz096}
  {\path{doi:10.1093/jamia/ocz096}}.

\bibitem{laparra2020RethinkingDomainAdaptation}
E.~Laparra, S.~Bethard, T.~A. Miller, Rethinking domain adaptation for machine
  learning over clinical language, JAMIA Open 3~(2) (2020) 146--150.
\newblock \href {http://dx.doi.org/10.1093/jamiaopen/ooaa010}
  {\path{doi:10.1093/jamiaopen/ooaa010}}.

\bibitem{smit2020CombiningAutomaticLabelersa}
A.~Smit, S.~Jain, P.~Rajpurkar, A.~Pareek, A.~Ng, M.~Lungren, Combining
  {{Automatic Labelers}} and {{Expert Annotations}} for {{Accurate Radiology
  Report Labeling Using BERT}}, in: Proceedings of the 2020 {{Conference}} on
  {{Empirical Methods}} in {{Natural Language Processing}} ({{EMNLP}}), 2020,
  pp. 1500--1519.
\newblock \href {http://dx.doi.org/10.18653/v1/2020.emnlp-main.117}
  {\path{doi:10.18653/v1/2020.emnlp-main.117}}.

\bibitem{fries2021OntologydrivenWeakSupervisionb}
J.~A. Fries, E.~Steinberg, S.~Khattar, S.~L. Fleming, J.~Posada, A.~Callahan,
  N.~H. Shah, Ontology-driven weak supervision for clinical entity
  classification in electronic health records, Nature Communications 12~(1)
  (2021) 2017.
\newblock \href {http://dx.doi.org/10.1038/s41467-021-22328-4}
  {\path{doi:10.1038/s41467-021-22328-4}}.

\end{thebibliography}

\end{document}